\documentclass{article}

\usepackage{PRIMEarxiv}

\usepackage{fontspec}       
\usepackage{hyperref}       
\usepackage{url}            
\usepackage{booktabs}       
\usepackage{amsfonts}       
\usepackage{amsmath}
\usepackage{nicefrac}       
\usepackage{microtype}      
\usepackage{graphicx}
\usepackage{array}
\usepackage[bidi=default]{babel}   
\babelprovide[import]{arabic}
\babelfont[arabic]{rm}[Path=fonts/,Script=Arabic]{DejaVuSans.ttf}
\newcommand{\textarabic}[1]{\foreignlanguage{arabic}{#1}}
\graphicspath{{media/}}     

\title{Language-Specific versus Cross-Lingual Knowledge Graphs for Implicit
Aspect Identification in Arabic: A Comparative Study of Reasoning and
Adaptation Strategies
}

\author{
  Lujain A. Alawwad \\
  Department of Computing and Informatics \\
  Saudi Electronic University \\
  Riyadh, Saudi Arabia \\
  \texttt{l.alawwad@seu.edu.sa} \\
}

\begin{document}
\maketitle

\begin{abstract}
Aspect-based sentiment analysis (ABSA) in Arabic must recover both explicitly
stated aspects and implicit aspects that are never named in the text. Implicit
identification typically relies on an auxiliary knowledge source (e.g., knowledge graph (KG)) linking opinion cues
to aspect categories, but for a lower-resource language the practitioner faces
a design choice: reuse a mature English KG through multilingual embeddings, or
build a smaller native Arabic KG. This paper reports a controlled comparison of
the two strategies within a single hybrid pipeline, evaluated on three Arabic
benchmarks (M-ABSA, SemEval-2016 Arabic, and HAAD). We further compare two
adaptation strategies for the generative extractor that feeds the
KG---zero-shot prompting versus task-specific fine-tuning of an 8B-parameter
large language model (LLM). The native Arabic KG
(Strategy~2) outperforms the cross-lingual English KG (Strategy~1) by
$+0.199$ micro-$F_1$ on M-ABSA and $+0.251$ on SemEval-2016, gaining on both
precision and recall. Task-specific fine-tuning raises explicit-extraction
micro-$F_1$ from $\leq 0.13$ (zero-shot) to $0.66$--$0.76$ on M-ABSA and
SemEval-2016 ($0.45$ on the smaller HAAD), confirming that task adaptation,
rather than model scale, is decisive in a morphologically rich language.
\end{abstract}


\section{Introduction}
Aspect-based sentiment analysis decomposes an opinionated sentence into the
specific targets being evaluated and the sentiment expressed toward each. A
recurring difficulty is the \textbf{implicit aspect}: a sentence such as
``too noisy to sleep'' expresses an opinion about \textit{comfort} without ever
naming it. Explicit-only systems, which label tokens that appear in the text,
cannot address such sentences by construction. Recovering implicit aspects
instead requires reasoning from opinion cues (adjectives, sentiment verbs) to
latent aspect categories, a mapping that is naturally encoded in a knowledge
graph.

For Arabic this raises a concrete engineering question. English benefits from large, well-curated aspect resources; Arabic does not. For constructing an Arabic implicit aspect module, there are two options: Either the team may leverage an already developed English knowledge graph and bridge the language gap using multilingual sentence embeddings, or they may develop a native Arabic knowledge graph and perform lexical matching directly. The former approach is cost-effective, while the latter involves higher costs but will prevent the loss of semantics due to cross-lingual projections.
Which approach prevails in practice is an empirical question that this paper addresses.

We report two comparisons, both on Arabic data and both within one hybrid
pipeline so that each factor is isolated: (1)~a \textbf{language-specific
Arabic KG versus a cross-lingual English KG} for implicit identification;
and (2)~\textbf{zero-shot prompting versus task-specific fine-tuning} of the
generative extractor that supplies explicit aspects and abstentions to the KG. The contributions are the head-to-head KG
comparison on identical Arabic test splits and the quantification of the
adaptation gap in a morphologically rich language. This investigation forms one component of a broader
study on implicit aspect extraction for Arabic and English; the full pipeline,
of which the knowledge-graph module examined here is a part, is presented in our
earlier work \cite{alawwad2026context}.

\section{Pipeline and Experimental Setup}
The evaluated system is a hybrid pipeline whose explicit-extraction backbone is
\textbf{Llama-3-8B} \cite{grattafiori2024llama3}, adapted per dataset with Quantized
Low-Rank Adaptation (QLoRA) \cite{hu2022lora} parameter-efficient fine-tuning.
The generative backbone produces surface aspect terms and, crucially, abstains
(predicts \texttt{NULL}) on sentences with no surface aspect---delegating those
to the KG. Two recall-oriented modules, iterative generation (mask-and-requery)
and paraphrase-augmented extraction, sit on top of the baseline. The
implicit-aspect module is the KG, which maps opinion cues to
Entity\#Aspect categories and is the focus of the first comparison below.

Three Arabic benchmarks are used (Table~\ref{tab:datasets}): \textbf{M-ABSA}
(Arabic, multi-domain: hotel, restaurant, food, laptop, phone, course, sight),
\textbf{SemEval-2016 Arabic} (SE-16, hotel reviews), and \textbf{HAAD} (Arabic
book reviews). Explicit extraction is measured by micro-averaged precision,
recall, and $F_1$ on explicit rows; implicit identification is evaluated
separately against Entity\#Aspect gold annotations. HAAD is excluded from the
implicit comparison because its test split contains no implicit aspects.
Statistical significance uses paired per-example tests (bootstrap and
McNemar's test).

\begin{table}[htbp]
\caption{Arabic benchmarks used in the study (explicit-row test counts).
M-ABSA additionally contains implicit and mixed rows evaluated separately.}
\centering
\renewcommand{\arraystretch}{1.5}
\begin{tabular}{llrl}
\toprule
\textbf{Dataset} & \textbf{Domain} & \textbf{Explicit rows (N)} & \textbf{Implicit gold} \\
\midrule
M-ABSA (AR)       & Multi-domain & 1,356 & Yes (718 implicit, 104 mixed) \\
SemEval-2016 (AR) & Hotel        & 1,130 & Yes (96 implicit) \\
HAAD (AR)         & Book reviews & 299   & None (explicit only) \\
\bottomrule
\end{tabular}
\label{tab:datasets}
\end{table}

\section{Language-Specific vs. Cross-Lingual Knowledge Graph}
The central comparison contrasts two constructions of the implicit-aspect KG,
evaluated against Entity\#Aspect gold annotations independently of explicit
extraction.

\subsection{Strategy 1 --- Cross-Lingual Semantic Matching}
\begin{figure}[htbp]
\centering
\includegraphics[width=0.42\linewidth]{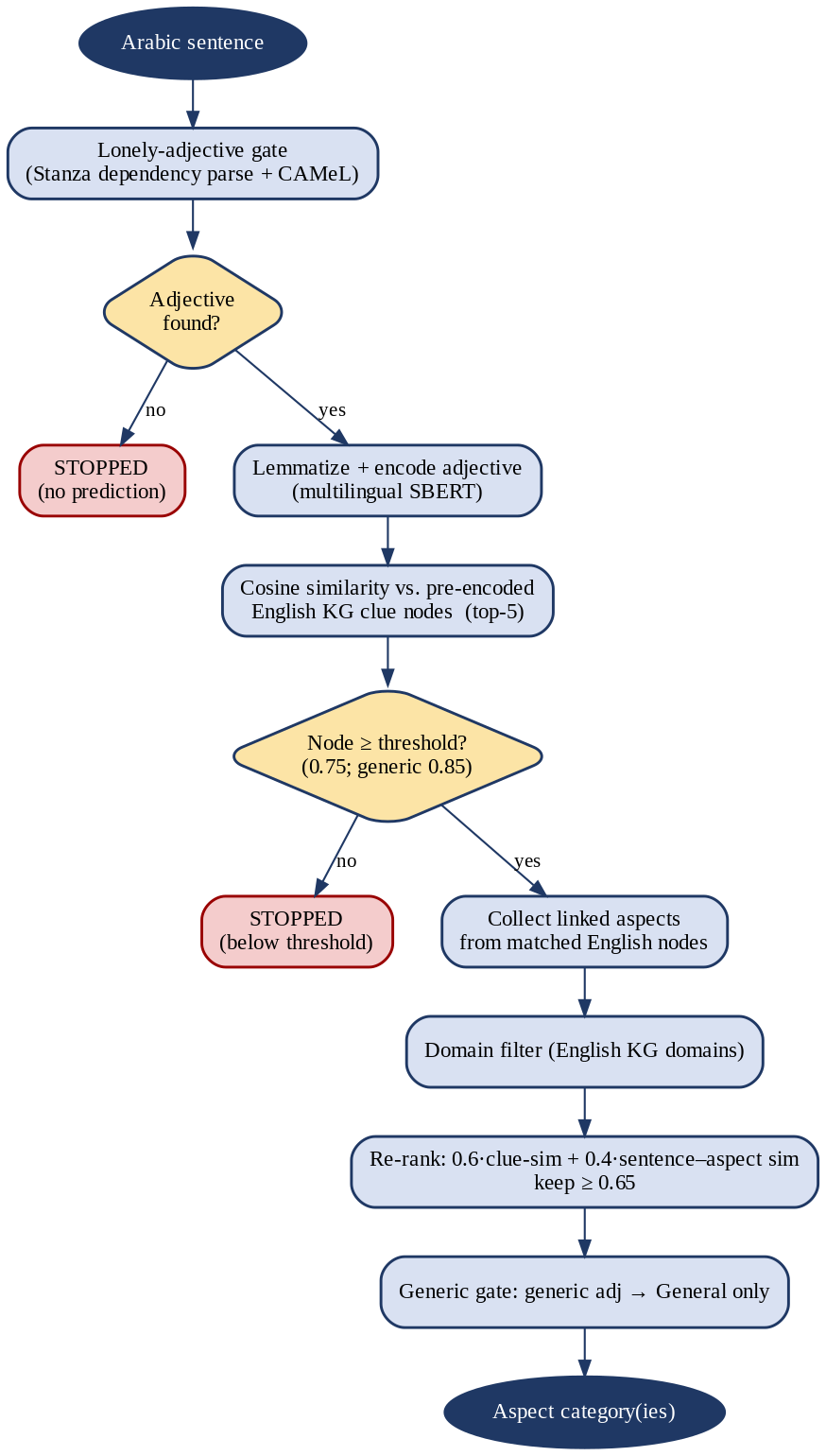}
\caption{Strategy~1 (cross-lingual semantic matching): Arabic opinion
adjectives are embedded and matched to English KG clue nodes by cosine
similarity, then filtered and re-ranked to aspect categories.}
\label{fig:s1}
\end{figure}

Strategy~1 reuses the mature English KG unchanged and bridges the language gap
with meaning rather than spelling. Each Arabic sentence first passes a
lonely-adjective gate---a Stanza dependency parse combined with CAMeL
morphological disambiguation isolates opinion adjectives that modify no explicit
aspect noun. Every such adjective is lemmatized and encoded with a multilingual
sentence encoder (\texttt{paraphrase-multilingual-MiniLM-L12-v2})
\cite{reimers2019sbert,reimers2020multilingual} into the same vector space as
the pre-encoded English clue nodes of the KG. The system retrieves the five
nearest English nodes by cosine similarity and keeps those above a similarity
threshold ($0.75$ for ordinary adjectives, raised to $0.85$ for generic
sentiment words such as ``good'' so they do not match everything). The aspects
linked to the surviving nodes are filtered to the sentence's inferred domain and
then re-ranked by a combined score that mixes the clue-to-node similarity
(weight $0.6$) with a direct sentence-to-aspect similarity (weight $0.4$);
aspects scoring below $0.65$ are discarded, and generic adjectives are
constrained to the \textsc{General} category. Semantic similarity guides
every step---matching, aspect selection, and ranking---so no exact Arabic
string ever has to appear in the graph. Figure~\ref{fig:s1} summarizes this
flow.

\subsection{Strategy 2 --- Native Arabic Lexical Matching}
\begin{figure}[htbp]
\centering
\includegraphics[width=0.5\linewidth]{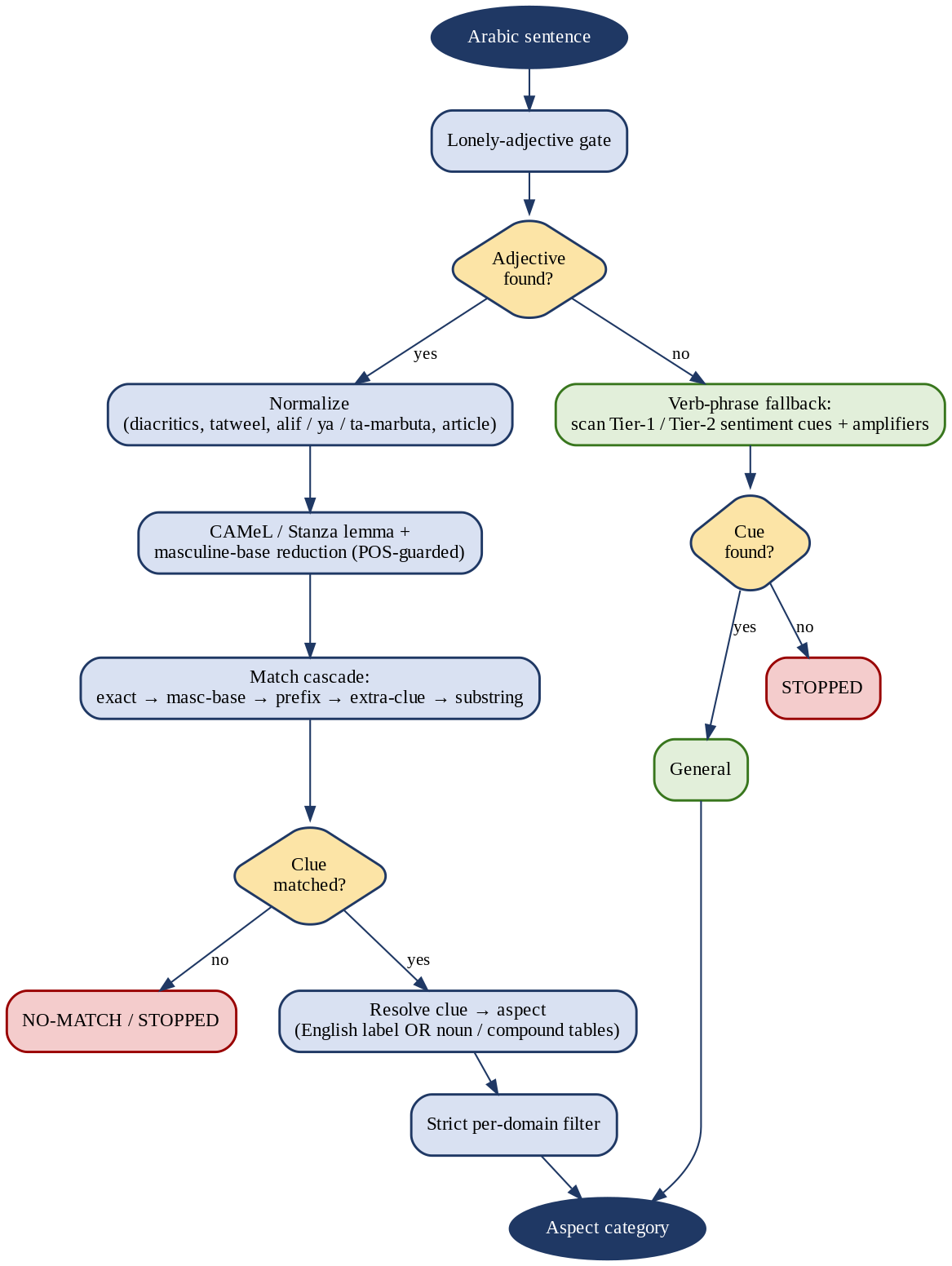}
\caption{Strategy~2 (native Arabic lexical matching): adjectives are
normalized, lemmatized, and reduced to a masculine base before an exact-match
cascade against the Arabic KG; sentences with no adjective take the verb-phrase
fallback path.}
\label{fig:s2}
\end{figure}

Strategy~2 builds a dedicated Arabic KG and matches on the surface form
directly, which demands a substantial normalization and morphology stack to make
the match succeed. After the same lonely-adjective gate, each candidate is
reduced through a cascade: full orthographic normalization (removing diacritics
and tatweel; unifying the hamza-carrying alif variants
\textarabic{أ، إ، آ} $\rightarrow$ \textarabic{ا}, \textarabic{ى} $\rightarrow$
\textarabic{ي}, and \textarabic{ة} $\rightarrow$ \textarabic{ه}; stripping the
definite article \textarabic{ال}), CAMeL and Stanza lemmatization, and a
feminine-to-masculine base reduction that strips gender, number, and case
suffixes while a CAMeL part-of-speech guard prevents it from truncating a
root-final ta marbuta in genuine nouns. The reduced forms are matched against
the KG clue set through an ordered cascade---exact, exact on the masculine base,
prefix, a curated extra-clue table, and finally substring. Matched clue nodes
resolve to aspect categories either directly (English-labeled edges) or through
two hand-built lookup tables that map Arabic nouns and adjective-noun compounds
(e.g., \textarabic{سهلة الوصول}, ``easy of access'' $\rightarrow$
\textsc{Location}) to categories, after which a strict per-domain filter is
applied. When no lonely adjective is found at all, a \textbf{verb-phrase
fallback} scans for tiered sentiment expressions: strong Tier-1 cues
(recommendation \textarabic{أوصي}/\textarabic{أنصح}, affect
\textarabic{أعجبني}/\textarabic{أحبه}, disappointment, and negation such as
\textarabic{لا يوجد} or \textarabic{تجنب}) and weaker Tier-2 cues
(\textarabic{سأعود}, ``I will return''; \textarabic{يستحق}, ``worth it'';
comparatives such as \textarabic{أرخص}, ``cheaper'' $\rightarrow$ \textsc{Prices}),
with amplifiers like \textarabic{جدا} promoting a weak cue to a strong one. This
fallback recovers the recommendation-heavy hotel sentences that carry sentiment
without any adjective. Figure~\ref{fig:s2} depicts the two matching paths.

\subsection{Design trade-offs}
The two designs trade coverage against precision in opposite directions.
Strategy~1 is cheap to build and inherently robust to morphological and
orthographic variation, because embeddings compare meaning: an inflection or
spelling never seen in the graph can still match its English counterpart. Its
weaknesses are the flip side of that abstraction---cross-lingual embeddings
occasionally align an Arabic cue to a plausible but wrong English node, the
aspect inventory and domains are inherited from English and may not fit Arabic
reviews, and the two thresholds plus the score weighting must be tuned. The net
effect is high precision but very low recall ($R = 0.10$--$0.14$): many Arabic
cues fall below threshold or are removed by the domain filter.

Strategy~2 is the opposite. Because it matches Arabic surface forms directly
against a native lexicon, it is precise and fully interpretable, and it natively
captures the Arabic sentiment verbs and compounds that embeddings miss---hence
its higher $F_1$. But it is brittle and preprocessing-heavy:
correctness depends on the normalization, lemmatization, and masculine-base
pipeline mapping the written word onto exactly the form stored in the clue node,
and the smallest uncovered morphological variation---an unexpected clitic, an
irregular plural, a dialectal spelling---fails to match, and the row is silently
dropped. It also requires the manually curated noun-to-aspect and verb-phrase
tables, which are labor-intensive to build and extend. This brittleness is the
direct cause of the high \texttt{STOPPED} rates in Table~\ref{tab:steps}:
Strategy~2 wins, but only because that manual effort was invested, and its
recall ceiling is a coverage problem inherent to exact lexical matching.
Table~\ref{tab:proscons} summarizes these trade-offs.

\begin{table}[htbp]
\caption{Design trade-offs between the two knowledge-graph strategies.}
\centering
\renewcommand{\arraystretch}{1.5}
\begin{tabular}{p{2.6cm}p{5.2cm}p{5.5cm}}
\toprule
\textbf{Dimension} & \textbf{Strategy 1 (Cross-lingual)} & \textbf{Strategy 2 (Native Arabic KG)} \\
\midrule
Matching basis        & Semantic similarity 
& Exact lexical lookup after normalization \\
Arabic resource       & None --- reuses the English KG & Dedicated Arabic KG + lookup tables \\
Morphology robustness & High (meaning-based) & Low (surface-based)\\
Preprocessing cost    & Light & Heavy 
\\
Interpretability      & Lower (embedding match) & High (explicit clue $\rightarrow$ aspect) \\
Precision / Recall    & High $P$, very low $R$ & Higher on both; recall capped by coverage \\
Main failure mode     & Wrong-domain node / below threshold & Uncovered morphology $\rightarrow$ no match (\texttt{STOPPED}) \\
\bottomrule
\end{tabular}
\label{tab:proscons}
\end{table}

\subsection{Quantitative Comparison}
Table~\ref{tab:s1s2} shows Strategy~2 is decisively better on both benchmarks.
On M-ABSA it improves micro-$F_1$ from $0.159$ to $0.358$ ($\Delta = +0.199$);
on SemEval-2016 from $0.217$ to $0.468$ ($\Delta = +0.251$). The gain is on
\textbf{both} precision and recall---cross-lingual projection loses on every
axis. Two mechanisms explain the advantage. First, strict native-domain
filtering removes cross-domain noise that multilingual similarity introduces
when an Arabic cue maps to a plausible but wrong English node. Second, the
Arabic clue lexicon with verb-phrase fallback recovers implicit rows that
embedding similarity alone misses, because recommendation and sentiment-verb
constructions are frequent in Arabic reviews but poorly aligned to English
adjective nodes.

\begin{table}[htbp]
\caption{Strategy~1 (cross-lingual English KG, multilingual embeddings) vs.
Strategy~2 (native Arabic KG, lexical lookup) for implicit identification.
Micro-averaged. Best $F_1$ per dataset in bold. HAAD excluded (no implicit
rows).}
\centering
\renewcommand{\arraystretch}{1.5}
\begin{tabular}{llrrr}
\toprule
\textbf{Dataset} & \textbf{Strategy} & \textbf{P} & \textbf{R} & \textbf{$F_1$} \\
\midrule
M-ABSA (AR)       & S1 cross-lingual & 0.484 & 0.095 & 0.159 \\
M-ABSA (AR)       & S2 Arabic KG     & 0.606 & 0.254 & \textbf{0.358} \\
SemEval-2016 (AR) & S1 cross-lingual & 0.455 & 0.143 & 0.217 \\
SemEval-2016 (AR) & S2 Arabic KG     & 0.698 & 0.352 & \textbf{0.468} \\
\bottomrule
\end{tabular}
\label{tab:s1s2}
\end{table}

Strategy~2's precision ($0.61$--$0.70$) is comparable to the English KG on
English data ($0.68$--$0.71$); the ceiling is recall. Table~\ref{tab:steps}
shows why: $52.1\%$ of M-ABSA and $42.7\%$ of SemEval-2016 implicit rows are
\texttt{STOPPED}---no cue is detectable---reflecting Arabic morphological
complexity and sentences that convey sentiment pragmatically rather than through
an adjective. The verb-phrase fallback is a substantial contributor: it recovers
$23.7\%$ of M-ABSA implicit rows, and inference fires on $53.1\%$ of
SemEval-2016 rows, mirroring the prevalence of recommendation expressions in
hotel reviews. Recall is therefore a \textbf{coverage} limitation with a clear
additive improvement path (extending the cue lexicon), not an architectural
ceiling.

\begin{table}[htbp]
\caption{Arabic KG (Strategy~2) inference-step distribution over implicit rows
(\%). \texttt{STOPPED} = no cue detected; \texttt{INFERRED} = category inferred
from a cue; \texttt{NO-MATCH} = cue found but no KG edge.}
\centering
\renewcommand{\arraystretch}{1.5}
\begin{tabular}{lrrr}
\toprule
\textbf{Dataset} & \textbf{STOPPED} & \textbf{INFERRED} & \textbf{NO-MATCH} \\
\midrule
M-ABSA (AR)       & 52.1 & 45.1 & 2.8 \\
SemEval-2016 (AR) & 42.7 & 53.1 & 4.2 \\
\bottomrule
\end{tabular}
\label{tab:steps}
\end{table}

\section{Zero-Shot Prompting vs. Task-Specific Adaptation}
The KG only receives the sentences on which the generative extractor abstains,
so the quality of the explicit extractor directly bounds the whole system. We
therefore compare two adaptation strategies for that extractor: \textbf{zero-shot
prompting} of Llama-3-8B (ZS) versus \textbf{task-specific LoRA fine-tuning} (BL,
the baseline). Table~\ref{tab:zsbl} reports explicit micro-$P$/$R$/$F_1$ on the
three Arabic benchmarks.

The gap is substantial. Zero-shot is near-unusable on Arabic ($F_1 \leq 0.13$ on
all three sets, and $\leq 0.05$ on M-ABSA and SemEval-2016), whereas
task-specific fine-tuning reaches $0.66$--$0.76$ on M-ABSA and SemEval-2016
($0.45$ on HAAD). Adaptation raises $F_1$ by roughly $0.61$ on M-ABSA and $0.72$
on SemEval-2016. The effect is larger than the analogous English gap,
underscoring that for a morphologically rich, comparatively lower-resource
language, \textbf{task adaptation---not model scale or prompt engineering---is
the single most decisive factor} for a generative formulation of aspect
extraction.

\begin{table}[htbp]
\caption{Zero-shot prompting (ZS) vs. task-specific LoRA fine-tuning (BL) of
Llama-3-8B, explicit extraction on Arabic datasets (micro-averaged). Best $F_1$
per dataset in bold.}
\centering
\renewcommand{\arraystretch}{1.5}
\begin{tabular}{lrrrrrr}
\toprule
& \multicolumn{3}{c}{\textbf{M-ABSA}} & \multicolumn{3}{c}{\textbf{SE-16}} \\
\cmidrule(lr){2-4}\cmidrule(lr){5-7}
\textbf{Setting} & \textbf{P} & \textbf{R} & \textbf{$F_1$} & \textbf{P} & \textbf{R} & \textbf{$F_1$} \\
\midrule
Zero-shot (ZS)  & 0.048 & 0.048 & 0.048 & 0.070 & 0.036 & 0.048 \\
Fine-tuned (BL) & 0.754 & 0.591 & \textbf{0.663} & 0.795 & 0.734 & \textbf{0.763} \\
\bottomrule
\end{tabular}
\label{tab:zsbl}
\end{table}

For completeness, HAAD follows the same pattern (ZS $F_1 = 0.125$ vs.
fine-tuned $0.454$). Beyond the baseline, the recall-oriented modules add value
conditionally: iterative generation lifts M-ABSA to $F_1 = 0.766$, and the full
pipeline (with paraphrasing) reaches $0.760$---a marginal decline consistent
with Arabic's sensitivity to paraphrase noise---while on SemEval-2016 the
baseline already saturates precision and paraphrasing is best kept minimal, a
divergence from English that a per-language configuration handles.

\section{Discussion}
Both comparisons converge on a single practical message for Arabic implicit ABSA. First, when building an implicit-aspect module, a native Arabic KG is worth the construction cost: it more than doubles micro-F1 over reusing an English KG through multilingual embeddings, gaining on both precision and recall.
However, the multilingual knowledge transfer shortcut while appealing, creates noise across domains and fails to take into account Arabic-specific sentiment-verb combinations. Finally, the generative extractor used for the knowledge graph is not usable without task-specific adaptation, while zero-shot prompting of the 8B model is nearly unusable on Arabic. 

The residual limitation is recall on implicit rows, driven by high
\texttt{STOPPED} rates on cue-free Arabic sentences. Because the verb-phrase
fallback already recovers a large share additively, extending the Arabic cue
lexicon is a concrete, low-risk path to closing the gap without architectural
change.

\section{Conclusion}
We compared language-specific versus cross-lingual knowledge graphs, and zero-shot
versus task-specific adaptation for Arabic aspect-based sentiment analysis, all
on identical Arabic test splits within one hybrid pipeline. A native Arabic KG outperforms a
cross-lingual English KG by $+0.199$ and $+0.251$ micro-$F_1$ on M-ABSA and
SemEval-2016; task-specific fine-tuning is the decisive adaptation factor,
lifting explicit $F_1$ from $\leq 0.13$ to $0.66$--$0.76$ ($0.45$ on HAAD).
For Arabic, language-specific lexical
resources and task adaptation both pay off, and the multilingual shortcut proves
counterproductive for implicit reasoning.

\bibliographystyle{unsrt}
\bibliography{references}

\begin{thebibliography}{1}

\bibitem{alawwad2026context}
Lujain~A. Alawwad and Mohamed El~Bachir Menai.
\newblock From context to aspects: {LLM}-based implicit aspect extraction with
  paraphrased input and knowledge graph support.
\newblock {\em AI}, 7(7):240, 2026.

\bibitem{grattafiori2024llama3}
Aaron Grattafiori and {others (Llama Team, Meta AI)}.
\newblock The {Llama}~3 herd of models.
\newblock {\em arXiv preprint arXiv:2407.21783}, 2024.

\bibitem{hu2022lora}
Edward~J. Hu, Yelong Shen, Phillip Wallis, Zeyuan Allen-Zhu, Yuanzhi Li, Shean
  Wang, Lu~Wang, and Weizhu Chen.
\newblock {LoRA}: Low-rank adaptation of large language models.
\newblock In {\em Proc. Int. Conf. Learning Representations (ICLR)}, 2022.

\bibitem{reimers2019sbert}
Nils Reimers and Iryna Gurevych.
\newblock Sentence-{BERT}: Sentence embeddings using siamese {BERT}-networks.
\newblock In {\em Proc. Conf. Empirical Methods in Natural Language Processing
  (EMNLP-IJCNLP)}, pages 3982--3992, 2019.

\bibitem{reimers2020multilingual}
Nils Reimers and Iryna Gurevych.
\newblock Making monolingual sentence embeddings multilingual using knowledge
  distillation.
\newblock In {\em Proc. Conf. Empirical Methods in Natural Language Processing
  (EMNLP)}, pages 4512--4525, 2020.

\end{thebibliography}

\end{document}